\documentclass[10pt,twocolumn,letterpaper]{article}

\usepackage{3dv}
\usepackage{times}
\usepackage{epsfig}
\usepackage{graphicx}
\usepackage{amsmath}
\usepackage{amssymb}
\usepackage{booktabs}
\usepackage{multirow}
\usepackage{caption}
\usepackage{subcaption}
\usepackage{algorithm}
\usepackage[noend]{algpseudocode}
\usepackage{xcolor}

\makeatletter
\def\BState{\State\hskip-\ALG@thistlm}
\makeatother

\usepackage{pifont}

\newcommand\sbullet[1][.5]{\mathbin{\vcenter{\hbox{\scalebox{#1}{$\bullet$}}}}}
\newcommand{\model}{Joint Language-to-Pose }
\newcommand{\modelshort}{JL2P}
\newcommand{\modelshorts}{JL2P }
\newcommand{\baseline}{Lin et. al.}
\newcommand{\baselines}{Lin et. al. }
\newcommand{\TITLE}{Language2Pose: Natural Language Grounded Pose Forecasting}

\usepackage[pagebackref=true,breaklinks=true,letterpaper=true,colorlinks,bookmarks=false]{hyperref}
\hypersetup{
    colorlinks=true,
    linkcolor=blue,
    filecolor=magenta,      
    urlcolor=cyan,
}

\threedvfinalcopy 

\def\threedvPaperID{230} 

\ifthreedvfinal\fi
\begin{document}

\title{\TITLE}

\author{Chaitanya Ahuja\\
Language Technologies Institute\\
Carnegie Mellon University\\
{\tt\small cahuja@andrew.cmu.edu}
\and
Louis-Philippe Morency\\
Language Technologies Institute\\
Carnegie Mellon University\\
{\tt\small morency@cs.cmu.edu}
}
\twocolumn[{%
\renewcommand\twocolumn[1][]{#1}%
\maketitle
\vspace{-25pt}
\begin{center}
\url{http://chahuja.com/language2pose}
\end{center} 
\begin{center}
    \centering
    \includegraphics[width=0.8\textwidth]{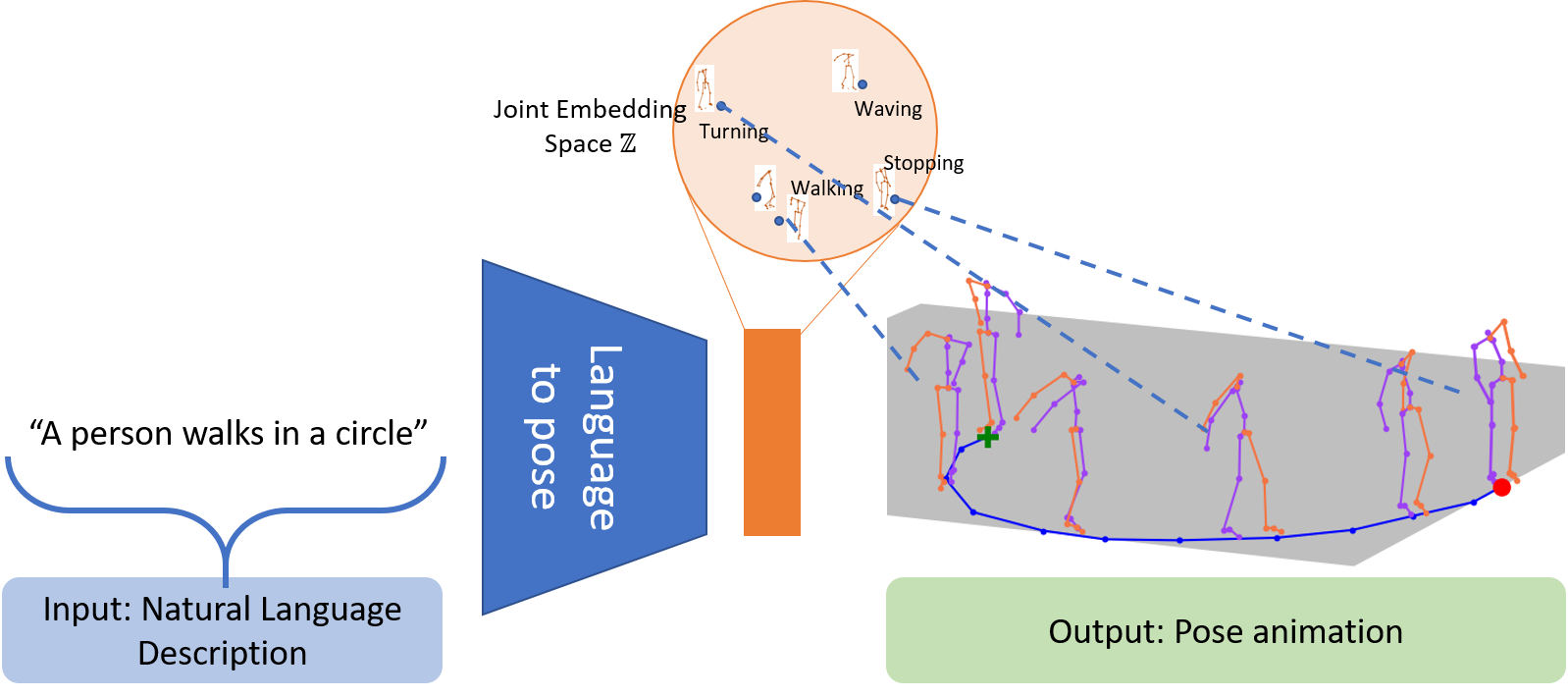}
    \captionof{figure}{Overview of our model which uses joint multimodal space of language and pose to generate an animation conditioned on the input sentence.}\label{fig:overview}
\end{center}%
}]
\pagenumbering{gobble}
\begin{abstract}
Generating animations from natural language sentences finds its applications in a a number of domains such as movie script visualization, virtual human animation and, robot motion planning. These sentences can describe different kinds of actions, speeds and direction of these actions, and possibly a target destination. The core modeling challenge in this language-to-pose application is how to map linguistic concepts to motion animations. 

In this paper, we address this multimodal problem by introducing a neural architecture called \model (or \modelshort), which learns a joint embedding of language and pose. This joint embedding space is learned end-to-end using a curriculum learning approach which emphasizes shorter and easier sequences first before moving to longer and harder ones. We evaluate our proposed model on a publicly available corpus of 3D pose data and human-annotated sentences. Both objective metrics and human judgment evaluation confirm that our proposed approach is able to generate more accurate animations and are deemed visually more representative by humans than other data driven approaches.
\end{abstract}

\section{Introduction}
\label{sec:intro}
\begin{figure*}
  \centering
  \includegraphics[width=\textwidth]{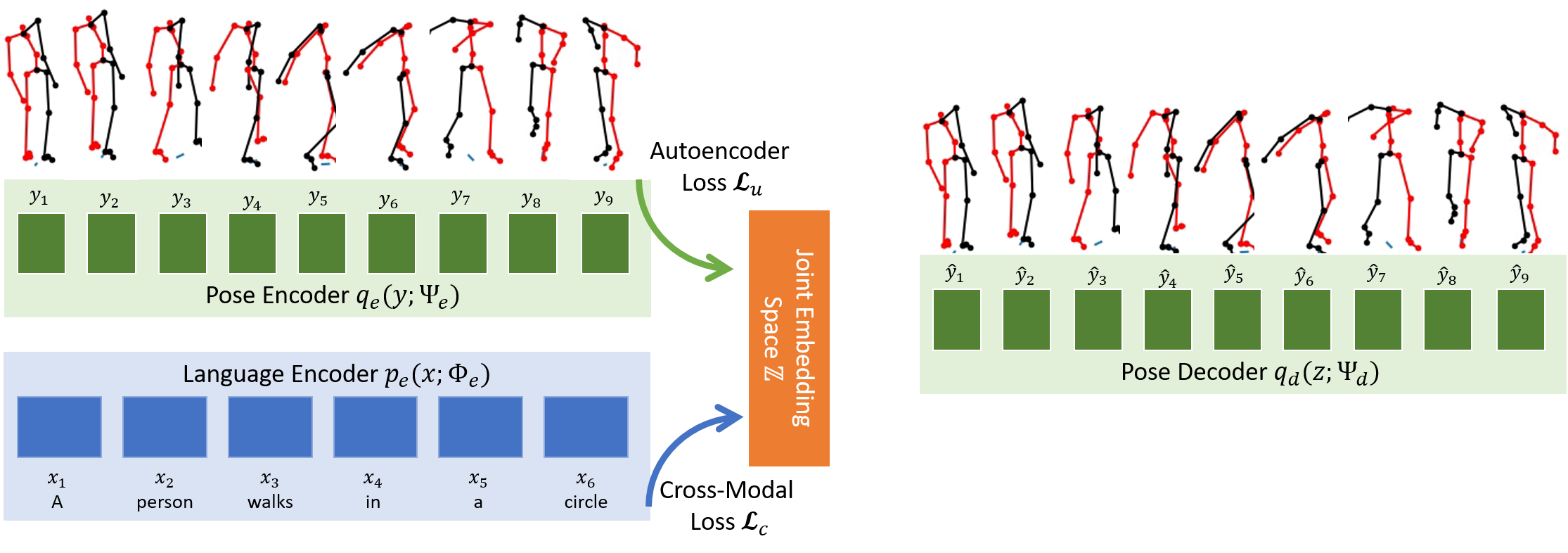}
  \caption{Overview of our proposed model \model (or \modelshort). Language and pose are mapped to a joint embedding space $\mathcal{Z}$, which can now be used by a trained pose decoder $q_d$ to generate a pose sequence. At train time both $p_e$ and $q_e$ are used to create the joint embedding using a training curriculum. But at inference time $z\in \mathcal{Z}$ is encoded by $p_e$ and decoded by $q_d$, giving us a model which can generate a animation (or sequence of poses) from a free form description (or language). }\label{fig:model}
\end{figure*} 
Generating animations from natural language descriptions is a first step for movie script visualization \cite{hanser2009scenemaker, ma2006virtual} which can later be stitched together while maintaining co-references in the story-line \cite{Zhang2019GeneratingAF}. These language grounded animations can also be useful in cases like virtual human animation \cite{takeuchi2017speech, chiu2015predicting, chiu2011train}, robot motion and task planning \cite{hwang1996interactive, ahn2018text2action}. 

An animation consists of a sequence of poses, which can be represented by positions of different joints in the body such as \emph{Root} (base of spine), \emph{head}, \emph{shoulder}, \emph{wrist}, \emph{knee} and many more. 

Pose forecasting conditioned on natural language has 3 major challenges. First, pose and natural language are very different modalities. The model needs a joint space where both natural language sentences and poses can be mapped. The model should also be able to decode animations from this embedding space. Second, different words of a sentence represent different qualities about the animation. Verbs and adverbs describe the action and speed/acceleration of the action; nouns and adjectives describe locations and directions respectively. The model has to map these concepts to small pose sequences and then stitch them to render convincing animations. Third, we want to see if objective metrics correlate with subjective metrics for this task as our models are trained using objective distance metrics, but the quality of generated animations can only be judged by humans.

In this paper, our two main contributions tackle the modeling challenges of pose and natural language. First, we propose a model \model (or \modelshort) that learns a joint embedding space of these two modalities. Second, we use a training curriculum to help the model emphasize more on shorter and and easier sequences first and longer and harder sequences later. Additionally, to make the training regimen robust to the outliers in the dataset, we use Smoooth L1 as the distant metric in our loss function. Through multiple objective and subjective experiments, we show that our model can generate more accurate and natural animations from natural language sentences than other data driven models.   



\section{Related Work}
\label{sec:related}
\begin{figure*}
  \centering
  \includegraphics[trim=100 20 100 50,clip,width=\textwidth]{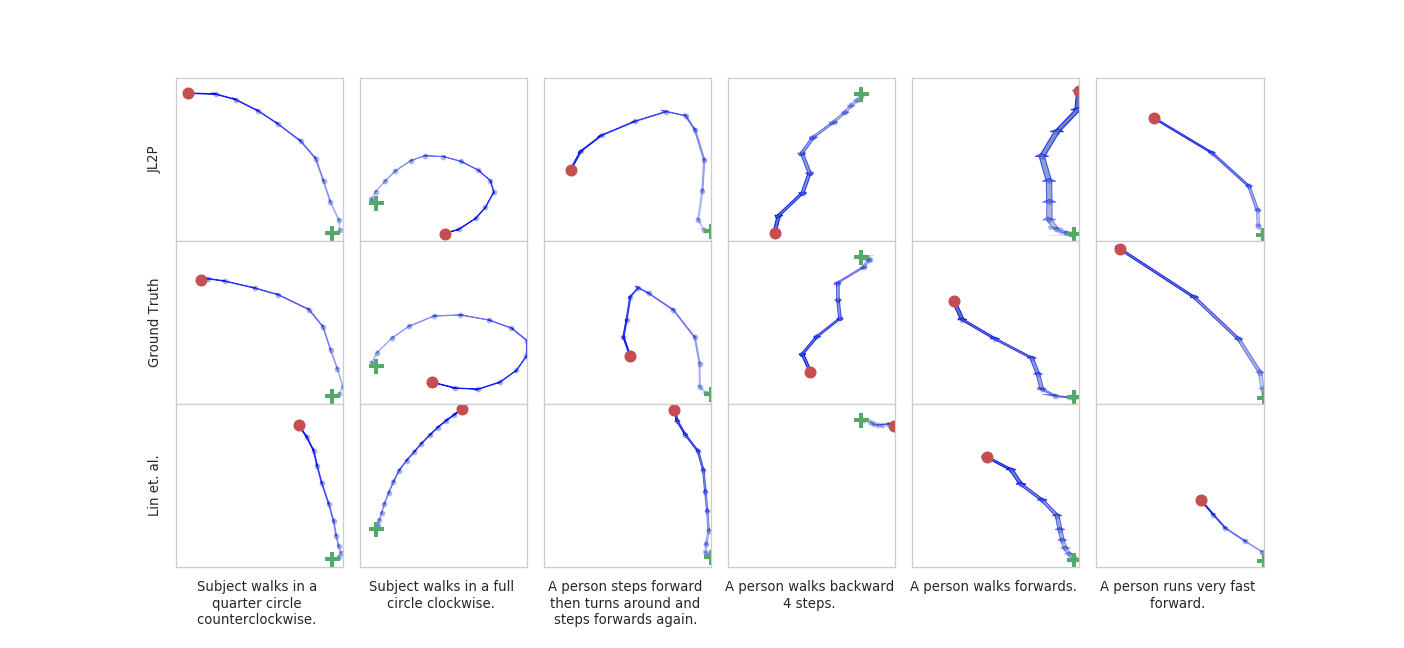}
  \caption{Trajectory plots of the generated pose (i.e. \emph{Root}'s position) viewed from the top. Each box represents a generated trajectory of the model on the vertical axis and sentence on the horizontal axis. The person starts at the green cross (\color{green}x\color{black}) and ends at the red circle (\color{red}$\bullet$\color{black}) with blue dots (\color{blue}$\sbullet$\color{black}) denoting equally placed time-steps. All trajectories in each column have the same scale for fair comparison across models.} \label{fig:trajectory}
\end{figure*}
\textbf{Pose Forecasting}: Data driven human pose forecasting attempts to understand the behaviours of the subject from its history of poses and generates the next sequence of poses. Short-term predictions \cite{pavllo2018quaternet} focus on modeling joint angles corresponding to hands, legs, head and torso. Long-term predictions \cite{ghosh2017learning, tang2018long, pavllo2018quaternet} additionally model the positions of the human character to generate animations like walking, running, jumping and crawling.

While some works use different actions (such as running, kicking, and more) as conditioning variables to generate the future pose \cite{tang2018long,lin2018human}, others rely solely on the history of poses to predict what kind of motion will follow \cite{chiu2019action}. Pose forecasting for locomotion is a more commonly researched topic, where models decide where and when to run/walk based on low-level control parameters such as trajectory and terrain \cite{holden2017phase}. Task based locomotion (such as writing on a whiteboard, moving a box, and sitting on a box) add the nuances of transitioning from one task to another, but pose generation is based on task-specific footstep plans that act as motion templates \cite{agrawal2016task}. 

All these approaches are either action specific, or require a set of low-level control parameters to forecast future pose. In this work, we aim replace low-level control parameters with high-level control parameters (e.g. natural language) to control actions and their speed and direction for the generated pose.

\textbf{Image or Speech conditioned pose forecasting}:
Images with a human can act as a context to forecast what comes next. Chao et. al. \cite{chao2017forecasting} use one image frame to predict the next few poses. These generated poses can now be used to aid the generation of a video \cite{yang2018pose} or a sequence of images \cite{ma2017pose}. An image, a high-level control parameter, has action information for pose generation, but it does not provide a fine-grained control on the speed and acceleration of the motion trajectory. 

Speech can also be used to control animations of virtual characters. Taylor et. al.\cite{taylor2017deep} use a data driven approach to model facial animation, while upper body pose forecasting conditioned on speech inputs has been tackled by Takeuchi et. al.\cite{takeuchi2017speech}. But, these pose sequences model the non-verbal behaviours (such as head nods, pose switches, hand waving and so on) of the character and do not offer fine-grained control over the characters next movements.

\textbf{Language conditioned pose forecasting}:
Natural language sentences consists of verbs describing the actions, adverbs describing the speed/acceleration of the action, and nouns with adjectives to describe the direction or target. This information can help provide a more fine-grained control over pose generations compared to image or speech.

Statistical models \cite{takano2015statistical, takano2012bigram} which use bigram models for natural language have been trained to encode motion sequences from sentences. Ahn et. al. \cite{ahn2018text2action} use around 2100 hours of youtube videos with annotated text descriptions to train a pose generation model. Pose sequences extracted from videos have limited translation and occluded lower bodies, hence their model only predicts the upper body with a static \emph{Root} joint. Some works use 3D motion capture data instead \cite{plappert2018learning, yamada2018paired}. 

Human motions generally have translation of the \emph{Root} joint, hence forecasting trajectory is important to get natural looking animations. Lin et. al \cite{lin20181} generates pose of all the joints of the body by pretraining a pose2pose autoencoder model before mapping language embeddings on the learned pose space. But the embedding space is not learned jointly \cite{pan2016jointly} which may limit the generative powers of the pose decoder. In contrast, our proposed approach learns a joint embedding space of language and pose using a curriculum learning training regime.



\section{Problem Statement}
\label{sec:background}
As an example, consider a natural language sentence which describes a human's motion: \emph{"A person walks in a circle"}. The goal of this cross-modal language-to-pose translation task is to generate an animation representing the sentence; i.e. an animation that shows a person following a trajectory of a circle with a walking motion (see figure \ref{fig:overview}).

Formally, given a sentence, represented by an N-sized sequence of words $X_{1:N} = \left[ x_1, x_2, \ldots x_N \right]$, we want to predict a T-sized sequence of 3D poses $Y_{1:T} = \left[ y_1, y_2, \ldots y_T\right]$ that are coherent with the semantics in the sentence. $x_i \in \mathcal{R}^K$ is the $i^{th}$ word vector with dimension $K$. $y_t \in \mathcal{R}^{J\times 3}$ is the pose matrix at time $t$. Rows of $y_t$ represent joints of the skeleton and columns are the $xyz$-coordinates of each joint. Tensors $X$ and $Y$ are elements of sets $\mathcal{X}$ and $\mathcal{Y}$ respectively.

Modeling language-to-pose is done by training a model $f: \mathcal{R}^{K\times N} \longrightarrow \mathcal{R}^{J\times 3 \times T}$ to predict a pose sequence $\hat{Y}_{1:T}$
\begin{equation}
    \hat{Y}_{1:T} = f \left( X_{1:N}; \Theta \right)
\end{equation}
where $\Theta$ are trainable parameters of the model $f$.

\section{\model}
\label{sec:model}
Language-to-pose models should be able to grasp nuanced concepts like speed, direction of motion and the kind of actions from the language and translate them to pose sequences (or animations). This requires the model to learn a multimodal joint space of language and pose. In doing so, it should also be able to generate sequences that are deemed correlated to the sentence by humans. 

To achieve that objective, we propose \model (or \modelshort) model to learn the joint embedding space. Given an input sentence, an animation can be sampled from this model at inference stage.

In this section, a joint embedding space of language and pose is formalized. This is followed by an algorithm to train for the joint embedding space and a discussion on the practical edge cases at inference time for our \model model.

\subsection{Joint Embedding Space for Language and Pose}
To learn a joint embedding space of language and pose, the sentence $X_{1:N}$ and pose $Y_{1:T}$ are first mapped to a latent representation using a sentence encoder $p_e(X_{1:N}; \Phi_e)$ and a pose encoder $q_e(Y_{1:T}; \Psi_e)$ respectively. These estimate the latent representation or embeddings $z_x$ and $z_y$ respectively in the embedding space $\mathcal{Z} \subset \mathcal{R}^h$, 

\begin{eqnarray}
z_x = p_e \left(X_{1:N}; \Phi_e \right)\\ \label{eq:encode}
z_y = q_e \left(Y_{1:T}; \Psi_e \right)
\end{eqnarray}

$z_x,z_y$ should lie close to each other in $\mathcal{Z}$ as they represent the same concept. To ensure that they do lie close together, a joint translation loss is constructed (refer to Figure \ref{fig:model}) and trained end to end with a training curriculum.

\subsection{Joint Loss Function}
\label{ssec:joint}
Once we have the embedding $z_x$ or $z_y$, a pose decoder $q_d(; \Psi_d)$ is used to generate an animation from the joint embedding space $\mathcal{Z}$. The output of the pose decoder must now lie close to the pose sequence $Y_{1:T}$. Hence, using $X_{1:N}$ as inputs and $Y_{1:T}$ as outputs, the cross-modal translation loss is defined as,
\begin{equation}
\label{eq:loss1}
    \mathcal{L}_c = d \left( q_d \left(  z_x; \Psi_d \right), Y_{1:T}\right)
\end{equation}
and using $Y_{1:T}$ as inputs and $Y_{1:T}$ as outputs, the uni-modal translation (or autoencoder) loss is defined as,
\begin{equation}
\label{eq:loss2}
    \mathcal{L}_u = d \left( q_d\left( z_y ; \Psi_d \right) , Y_{1:T}\right)
\end{equation}
where $d(x, y)$ is a function to calculate the distance between the predicted values and ground truth of pose. $\Phi_e$, $\Psi_e$ and $\Psi_d$ are trainable parameters of the sentence encoder, pose encoder and pose decoder respectively.

Combining equations \ref{eq:loss1} and \ref{eq:loss2} we get a joint translation loss,
\begin{equation}
    \label{eq:loss3}
    \mathcal{L}_j = \mathcal{L}_c + \mathcal{L}_u
\end{equation}
Jointly optimizing the loss $\mathcal{L}_j$ pushes $z_x$ and $z_y$ closer together improving generalizability and additionally trains the pose decoder which is useful for inference from the joint embedding space.

As $\mathcal{L}_j$ is a mutivariate function in $X_{1:N}$ and $Y_{1:T}$, coordinate descent \cite{wright2015coordinate} for optimizing the loss function is a natural choice and is described in Algorithm \ref{alg:1}.

\subsection{Training Curriculum}
\label{ssec:curriculum}
Cross modal pose forecasting can be a challenging task to train \cite{chao2017forecasting}. Starting with simpler examples before moving on to tougher ones can be beneficial to the training process \cite{bengio2009curriculum, zaremba2014learning, yang2015weakly}. 

The curriculum design commonly used for pose forecasting \cite{chao2017forecasting} is adapted for our joint model. We first optimize the model to predict 2 time steps conditioned on the complete sentence. This easy task helps the model learn very short pose sequences like leg motions for walking, hand motions for waving and torso motions for bending. Once the loss on the validation set starts increasing, we move on to the next stage in the curriculum. The model is now given twice the amount of poses for prediction. The complexity of the task is increased in every stage till the maximum time-steps ($T$) of prediction is reached. We describe the complete training process in Algorithm \ref{alg:1}.

\begin{algorithm}
\caption{Learning language-pose joint embedding}\label{alg:1}
\begin{algorithmic}[1]
\Procedure{Initialization}{}
\State $\mathcal{X}_{train} ,\mathcal{X}_{val}, \mathcal{Y}_{train}, \mathcal{Y}_{val} \gets \text{SplitData}(\mathcal{X}, \mathcal{Y})$
\State $\text{MaxValLoss} \gets \inf$
\State $t \gets 2$
\EndProcedure
\Procedure{Curriculum}{}
\While{$t \leq T$}
\ForAll{$X_{1:N}, Y_{1:t} \in \mathcal{X}_{train}, \mathcal{Y}_{train}$}
\State $r \gets \text{CoinFlip}()$ // For Coordinate Descent
\If{$r == 0$}
\State $z \gets p_e(X_{1:N}; \Phi_e)$  //Encoder
\Else
\State $z \gets q_e(Y_{1:t}; \Psi_e)$  //Encoder
\EndIf
\State $\hat{Y_{1:t}} \gets q_d(z; \Psi_d)$  //Decoder
\State $\text{loss} \gets d(Y_{1:t}, \hat{Y}_{1:t})$
\State $\Phi_e, \Psi_e, \Psi_d \gets \text{UpdateModelParams}(\text{loss})$ 
\EndFor
\State $\text{ValLoss} \gets CalcValLoss (\mathcal{X}_{val}, \mathcal{Y}_{val})$
\If{$\text{ValLoss} > \text{MaxValLoss}$}
\State $t \gets 2t$
\State $\text{MaxValLoss} \gets \inf$
\EndIf
\EndWhile
\EndProcedure
\end{algorithmic}
\end{algorithm}

\subsection{Optimization}
For the distance metric $d(x,y)$ in Equation \ref{eq:loss1}, \ref{eq:loss2} and \ref{eq:loss3}, Smooth L1 loss (similar to Huber Loss \cite{huber1992robust}) is used which is defined as,

\begin{equation}
    SmoothL1(x, y)=
    \begin{cases}
    0.5(x-y)^2 & \text{for } |x-y|<1 \\
    |x-y| - 0.5 & \text{otherwise}
    \end{cases}
\end{equation}
In contrast, Lin et. al.\cite{lin20181} uses L2 loss for $d(x,y)$. L2 loss is more sensitive to outliers than L1 loss due to its linearly proportional gradient with respect to the error, while L1 loss has a constant gradient of 1 or -1. But L1 Loss can become unstable when $|x-y| \approx 0$, due to oscillating gradients between 1 and -1. On the other hand, Smooth L1 is continuous and smooth near 0 and more generally for all $x, y \in \mathcal{R}$, hence it is more stable than L1 as a loss function.



\begin{table*}[]
\begin{tabular}{@{}l|cc|lcccccccc@{}}
\toprule
\multicolumn{1}{c|}{\multirow{2}{*}{\textbf{Models}}}                   & \multicolumn{11}{c|}{\textbf{Average Positional Error (APE) in mm}}                                                                                                                                                                  \\ \cmidrule(l){2-12} 
\multicolumn{1}{c|}{}                                                   & \textbf{Mean} & \textbf{\begin{tabular}[c]{@{}c@{}}Mean w/o\\ Root\end{tabular}} & \textbf{Root}  & \textbf{Torso} & \textbf{Head} & \textbf{LArm} & \textbf{RArm} & \textbf{LHip} & \textbf{RHip} & \textbf{LFoot} & \textbf{RFoot} \\ \midrule
\textbf{\baseline\cite{lin20181}}                                               & \multicolumn{1}{l}{54.9$^{***}$}          & 50.0                                                             & 151.6          & 26.6           & 35.4          & 61.3          & 61.6          & 32.2          & 32.1          & 63.3           & 63.2           \\ \midrule
\textbf{\begin{tabular}[c]{@{}l@{}}\modelshort \\ w/o Curriculum\end{tabular}} & \multicolumn{1}{l}{52.2$^{***}$}          & 47.9                                                             & 139.2          & 24.2           & 32.5          & 57.3          & 57.2          & 30.6          & 30.7          & 62.9           & 63.2           \\ \midrule
\textbf{\begin{tabular}[c]{@{}l@{}}\modelshort \\ w/o L1\end{tabular}}         & \multicolumn{1}{l}{51.7$^{**}$}          & 47.0                                                             & 145.0          & 24.4           & 32.8          & 58.0          & 57.6          & 29.9          & 30.7          & 59.3           & 59.8           \\ \midrule
\textbf{\begin{tabular}[c]{@{}l@{}}\modelshort \\ w/o Joint Emb.\end{tabular}} & \multicolumn{1}{l}{50.4}          & 45.7                                                             & 143.3          & 24.0           & \textbf{31.0} & 55.6          & \textbf{54.5} & 29.7          & 29.5          & \textbf{59.0}  & 59.5           \\ \midrule
\textbf{\modelshort}                                                           & \multicolumn{1}{l}{\textbf{49.5}} & \textbf{45.4}                                                    & \textbf{131.1} & \textbf{23.0}  & 31.4          & \textbf{55.3} & 55.0          & \textbf{28.6} & \textbf{29.0} & 59.2           & \textbf{58.8}  \\ \bottomrule
\end{tabular}
\caption{Average positional error (APE) for \modelshorts, \modelshorts w/o Joint Emb., \modelshorts w/o L1, \modelshorts w/o Curriculum and \baseline. Lower is better. Our models (\modelshorts and variants) show consistent increase in accuracy over \baselines across all joints with the addition of components joint embedding, smooth L1 loss and curriculum learning. Two-tailed pairwise t-test between all models and \modelshorts where $***$- p$<$0.001, and $**$- p$<$0.01.}\label{tab:main}
\end{table*}

\section{Experiments}
\label{sec:exp}
Joint language to pose modeling can be broken down into three core challenges,
\begin{enumerate}
    \item \textbf{Prediction Accuracy by Joint Space}: How accurate is pose prediction from the joint embedding ?
    \item \textbf{Human Judgment}: Which of the generated animation is more representative of the input sentence? Does the subjective evaluation correlate with the results from the objective evaluations? 
    \item \textbf{Modeling nuanced language concepts}: Is the model able to map nuanced concepts such as speed, direction and action in the generated animations?
\end{enumerate}
Experiments are designed to evaluate these challenges of language grounded pose forecasting.

In the following subsections, the dataset and its pre-processing is briefly discussed which is followed by the evaluation metrics for both objective and subjective evaluations. Finally, design choices of the encoder and decoder models are described which are used to construct the baselines in the final subsection.

\subsection{Dataset}
Our models are trained an evaluated on KIT Motion-Language Dataset \cite{Plappert2016} which combines human motion with natural language descriptions. It consists of 3911 recordings (approximately 11.23 hours) which are re-targeted to a kinematic model of a human skeleton with 50 DoFs (6 DoF for the \emph{Root} joint's orientation and position, while remaining 44 DoFs for arms legs, head and torso). The dataset also consists of 6278 English sentences (approximately 8 words per sentence) describing the recordings. This is more than the number of recordings as each recording has one or more descriptions which are annotated by human volunteers. We use 20\% of the data as a randomly sampled held-out set  for evaluating all models.

There is wide variety of motions in this dataset ranging from locomotion (e.g. walking, running, jogging), performing (e.g. playing violin/guitar), and gesticulation (e.g. waving). Many recording have adjectives to further describe the motion like speed (e.g. fast and slow), direction (left, right and forward), and number for periodic motions (e.g. walk 4 steps). 

We use the pre-processing steps used in Holden et. al. \cite{holden2016deep}. All the frames of the motion are transformed such that body always faces the Z-axis. Joint rotation angles are transformed to 3D positions is the skeleton's local frame of reference with \emph{Root} as the origin. \emph{Root}'s position on XZ-plane and orientation along Y-axis is represented as velocity instead of absolute values. 

Motion sequences are then sub-sampled to a frequency of 12.5 Hz down from 100Hz. This is low enough to bring enough variance between 2 time-steps for the decoder to train for a regression task, while not compromising on the human's perception of the animation \cite{chao2017forecasting}. 

\subsection{Implementation Details}

For pose encoder ($q_e$) a network of Gated Recurrent Units (GRUs) \cite{cho2014properties} is used in our model \modelshort. The pose decoder ($q_d$) is the same except it has residual connection from the input to the output layer. This is similar to the pose decoder in Lin et. al. \cite{lin20181}, except an extra layer to predict the trajectory (or Trajectory Predictor) is discarded in our model. 

For langauge encoder ($p_e$), a network of Long-Short Term Memory Units (LSTMs) \cite{hochreiter1997long} is used. Each token of the sentence is converted into a distributional embedding using a pre-trained Word2Vec model \cite{mikolov2013distributed}. \footnote{We also train a variant of the model with BERT as the language encoder, but it did not show any significant improvements.} 


\subsection{Baselines}
There has been limited work done in the domain of data-driven cross-modal translation from natural language descriptions to pose sequence generation. The closest work to our proposed approach is by \textbf{\baselines} \cite{lin20181}\footnote{As we could not find code or pre-trained models for this work, we use our own implementation and training on the same data as all other baselines}. As mentioned in Section \ref{sec:related}, their model does not follow a training curriculum and uses L2 loss as the loss function. Thier model also maps the language embeddings to an existing embedding space of poses instead of jointly learning it.

We also compare our model \textbf{\modelshorts} (see Section \ref{sec:model}) with three ablations derived from itself. These ablations study the 3 main components of the model, joint embedding space, curriculum learning and Smooth L1 loss:
\begin{itemize}
    \item \textbf{\modelshorts w/o Curriculum} - Training curriculum in Section \ref{ssec:curriculum} is dropped. 
    \item \textbf{\modelshorts w/o L1} - L2 loss is used instead of Smooth L1 loss as the distance metric $d(x,y)$.
    \item \textbf{\modelshorts w/o Joint Emb.} - Instead of joint training as described in Section \ref{ssec:joint}, autoencoder loss $\mathcal{L}_u$ minimized first followed by optimization of the cross-translation loss $\mathcal{L}_c$.
\end{itemize}
\begin{figure*}
  \centering
  \includegraphics[width=\textwidth]{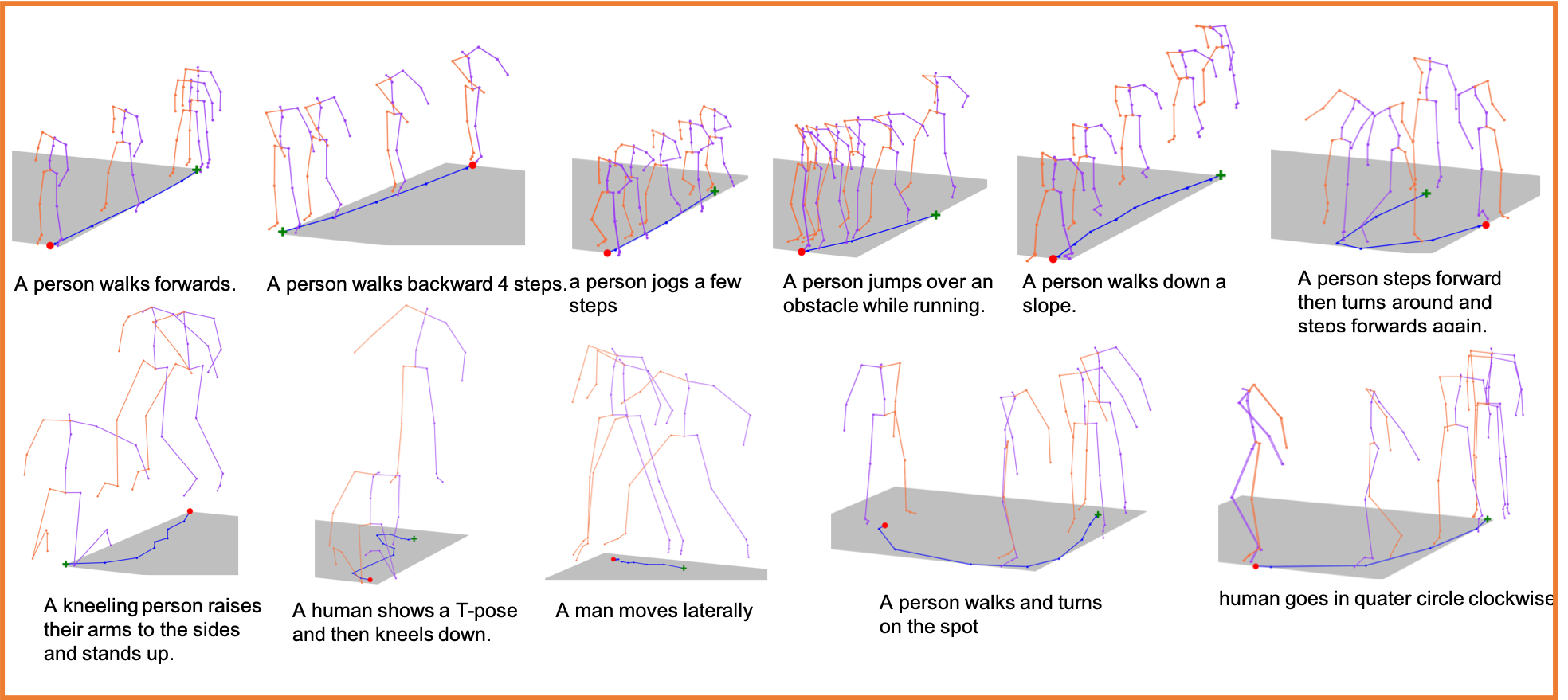}
  \caption{Renders of generated animations with a diverse set of sentences as input by our proposed model. Our model is able to change speed, direction and actions based on changes in the input sentence. Trajectory of the character is drawn with a blue line which starts at the green cross (\color{green}x\color{black}) and ends at the red circle (\color{red}$\bullet$\color{black}). }\label{fig:renders}
\end{figure*}

\subsection{Objective Evaluation Metrics}
\label{ssec:objective}
All models are evaluated on the held-out set with a metric Average Position Error (APE). Given a particular joint j, it can be denoted as $\mbox{APE}(j)$, 
\begin{equation}
    \mbox{APE}(p) = \frac{1}{\mathcal{Y}}\sum_{\mathcal{Y}} \|\hat{y}_t[j] - y_t[j] \|_2 
\end{equation}
where $y_t[j]$ is the true location and $\hat{y}_t[j] \in \mathcal{Y}$ is the predicted location of joint $j$ at time $t$

Another metric, Probability of Correct Keypoints (PCK) \cite{andriluka20142d, simon2017hand}, is also used as an evaluation metric. 

\subsection{User Study: Subjective Evaluation Metric}
\label{ssec:subjective}
Joint language to pose generation is a subjective task, hence a human's subjective judgment on the quality of prediction is an important metric for this task. 

To achieve this, we design a user study which asked human annotators to rank two videos generated by 2 different models but with same sentence as the input. One of the videos is generated by \baselines and the other is either ground truth or generated by \modelshorts, \modelshorts w/o Curriculum, \modelshorts w/o Joint Emb., or \modelshorts L1.
The annotators answer the following question for each pair of videos, \emph{Which of the 2 generated animations is better described by "$<$sentence$>$"?}. To ensure that annotators spend enough time to decide, any annotations which took less than 20 seconds\footnote{each video is 8 seconds long at an average. We set a threshold of 20 seconds to give annotators 4 seconds to make their decision.} were rejected. This study subjectively evaluates the preference of humans for generated animations by different models.

\section{Results and Discussion}
\label{sec:results}

In this section we first use objective measures and then conduct a user study to get a subjective evaluation. Finally, we probe some qualitative examples to understand the effectiveness of the model in tackling the core challenges described in Section \ref{sec:exp}.


\subsection{Prediction Accuracy by Joint Space}
\modelshorts demonstrates at least a 9\% improvement over \baselines (see Table \ref{tab:main}) for all joints. The maximum improvement around 15\% is seen in the \emph{Root} joint. Errors in \emph{Root} prediction can lead to a "sliding-effect" of the feet; when the generation is translating faster than the frequency of the feet. Improvements in APE scores for long-term prediction, especially for \emph{Root}, can help get rid of these artifacts in the generated animation.

When compared to its variants, \modelshorts loses maximum APE value when it is trained without curriculum (or \modelshorts w/o Curriculum). As discussed in Section \ref{ssec:curriculum}, learning to predict shorter sequences before moving on to longer ones proves beneficial for pose generation. APE scores go down by 4\%, if L2 loss is used instead of Smooth L1. In an output space as diverse as pose sequences, it becomes important to ignore outliers which may drive model to overfit.
APE scores go down only by 1\%, if the embedding space is not trained with the joint loss $\mathcal{L}_j$

\begin{figure}[!t]
  \centering
  \includegraphics[width=0.5\textwidth]{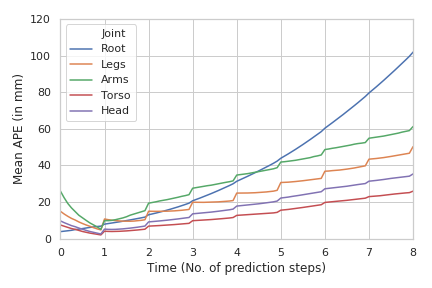}
  \caption{Plot of mean APE values across time for different parts of the body (\emph{Root, Legs, Arms, Torso} and \emph{head}) for \modelshort. Lower is better. Generating trajectory of the animating character is harder than the other joints as \emph{Root}'s APE blows up after around 500ms}\label{fig:ape_joint}
\end{figure}

APE values across time for \modelshorts of different parts of the body (\emph{Root, Legs, Arms, Torso} and \emph{Head}) are plotted in Figure \ref{fig:ape_joint}. \emph{Root}'s APE scores have the fastest rate of increase, followed by \emph{Arms, Legs} and then \emph{Head, Torso}. Two out of three coordinates of \emph{Root} are represented as velocity which accumulates errors when integrated back to absolute positions; this is probably a contributing factor to the rapid increase of prediction error over time.

\begin{figure}[!t]
  \centering
  \includegraphics[width=0.5\textwidth]{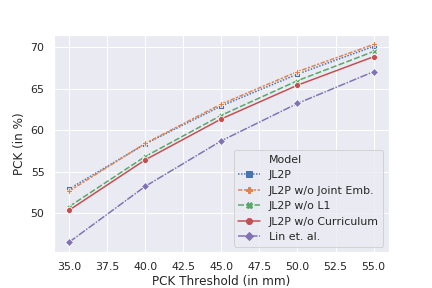}
  \caption{Plots of average Probability of Correct Keypoint (PCK) values over multiple values of thresholds ($\sigma$) for \modelshorts, \modelshorts w/o Joint Emb., \modelshorts w/o L1, \modelshorts w/o Curriculum and \baseline. Our model \modelshorts shows consistent improvements over other baselines across a large range of thresholds. Higher values are better.}
  \label{fig:pck}
\end{figure}

Our final objective metric is PCK. PCK values (for 35$\leq\sigma\leq$55) on generated animations are compared among \modelshorts, its variants and \baselines in Figure \ref{fig:pck}. \modelshorts and its ablations show a consistent improvement over \baselines which further strengthen the claim about the prediction accuracy by our model's joint space.
\begin{figure}[!t]
  \centering
  \includegraphics[width=0.5\textwidth]{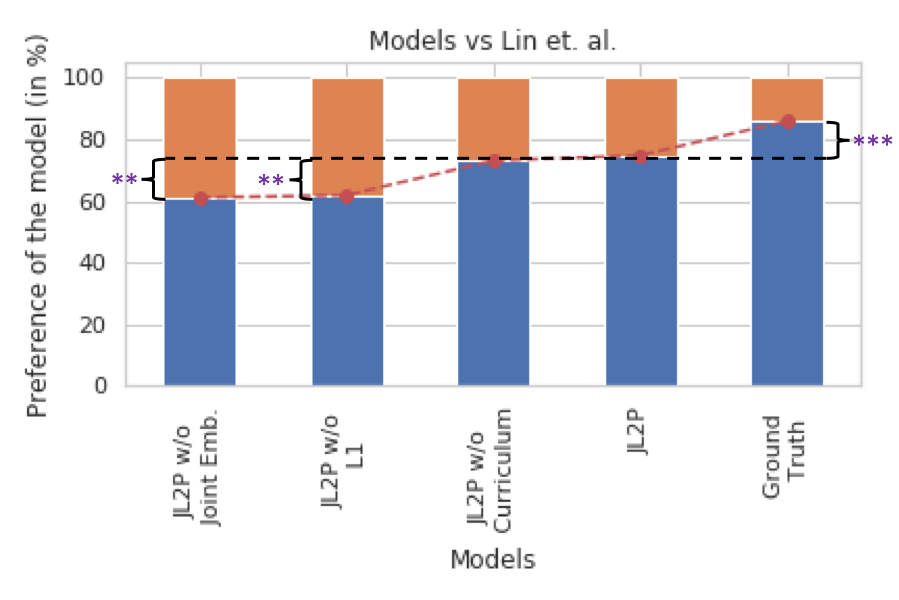}
  \caption{Preference scores of baseline models vs \baseline. Blue bars denote the preference percentage of models marked on the horizontal axis. Our models (\modelshorts and variants) show consistent rise in preference over \baselines with the addition of components joint embedding, smooth L1 loss and curriculum learning. **- $p < 0.01$ and ***- $p < 0.001$ for McNemar's test \cite{mcnemar1947note} on paired nominal data. }\label{fig:userstudy}
\end{figure}

\subsection{Human Judgment}
Human judgment is quantified by preference scores in Figure \ref{fig:userstudy}. Human preference of all our baseline models and ground truth are compared against \baselines animations. \modelshorts has a preference of 75\% which is shy of ground truth by 10\%. Preference scores consistently drop with all the other variants of \modelshorts. 

\modelshorts w/o Joint Emb. has the lowest preference score of 60\% when ranked against \baselines. It is still more preferred than \baselines but far more unlikely to be picked when pitted against \modelshorts. This is an interesting change in trend, as removing joint loss from \modelshorts did not affect the objective scores significantly, but have lowered its human preference by a significant fraction. This leads us to conclude that objective metrics are not enough to judge the performance of a model. Instead a combination of human judgment and objective metrics is necessary for evaluating pose generation models.

\subsection{Modeling nuanced language concepts}
\emph{Root} joint decides the trajectory of the animation which is crucial for translating concepts like speed (e.g. fast, and slow), direction (e.g. left, right, forward and backward) from natural language to animation. We plot the trajectories generated by \modelshort, ground truth and \baselines for different sentences in Figure \ref{fig:trajectory}.

\textbf{Modeling direction:} Animations' trajectory for these sentences for \modelshorts is similar to that of the ground truth trajectories. In contrast, \baseline's trajectories tend to be semantically incorrect and have a slightly curved forward motion for these sentences.

\textbf{Modeling speed:} In the sentence, "A person runs very fast forward", \modelshorts is able to understand that the animation has to move faster. It is able to walk approximately the same distance as the ground truth in the same amount of time, hence has the same speed. In contrast, even though \baseline's motion is in the forward direction, it is not able to maintain the same speed as required by the sentence. 

\textbf{Modeling actions: } In figure \ref{fig:renders}, we plot animations generated by a diverse set of sentences. \modelshorts is able to understand the action from the sentences, and is able to generate an animation corresponding to the action. \modelshorts is able to handle many actions ranging from \emph{kneeling} (with complex leg motions) to \emph{jogging} (with periodic hand and leg motion).

We show, via qualitative examples, that our model \modelshorts is able to model nuanced language concepts which are then reproduced in the animations generated at inference time.








\section{Conclusions}
\label{sec:conclusions}
In this paper, we proposed a neural architecture called \model (or \modelshort), which integrates language and pose to learn a joint embedding space in an end-to-end training paradigm. This embedding space can now be used to generate animations conditioned on an input description. We also proposed the use of curriculum learning approach which forces the model to generate shorter sequences before moving on to longer ones. We evaluated our proposed model on a parallel corpus of 3D pose data and human-annotated sentences with objective metrics to measure prediction accuracy, as well as with a user study to measure human judgment. Our results confirm that our approach, to learn a joint embedding in a curriculum learning paradigm by \modelshort, was able to generate more accurate animations and are deemed more visually represented by humans than the state-of-the-art model.

\section{Acknowledgements}
This material is based upon work partially supported by the National Science Foundation (Award \#1750439) and Oculus VR. Any opinions, findings, conclusions or recommendations expressed in this material are those of the author(s) and do not necessarily reflect the views of National Science Foundation or Oculus VR. No official endorsement should be inferred.

{\small
\bibliographystyle{ieee}
\bibliography{egbib}
}

\end{document}